\documentclass{vgtc}                          % final (conference style)
\graphicspath{{figures/}{pictures/}{images/}{./}} % where to search for the images

\usepackage{times}                     % we use Times as the main font
\usepackage{tabu}                      % only used for the table example
\usepackage{booktabs}                  % only used for the table example
\usepackage{lipsum}                    % used to generate placeholder text
\usepackage{mwe}                       % used to generate placeholder figures
\usepackage{algorithm}
\usepackage{algpseudocode}
\usepackage{mathptmx}                  % use matching math font
\usepackage{bm}                       % for bold math symbols
\usepackage{amsmath}                  % for math environments and symbols
\usepackage{textcomp}                 % for \textdegree
\usepackage{subcaption}
\usepackage{titlesec}
\titlespacing*{\section}{0pt}{0.35em}{0.15em}
\titlespacing*{\subsection}
{0pt}{0.45em plus 0.1em minus 0.1em}{0.20em}
\titlespacing*{\paragraph}{0pt}{0.20em}{0.35em}

\onlineid{0}

\vgtccategory{Research}

\vgtcinsertpkg

\title{AgenticFocus: Object-Preserving Mixed Reality Synthesis from Human FPV Video for Dexterous Humanoid Learning}

\newcommand{\equalcontribsymbol}{\ensuremath{{}^{*}}}
\newcommand{\equalcontrib}{\nobreak\hspace{0.35em}\equalcontribsymbol}
\author{
Iaroslav Kolomiets$^{1,2}$\thanks{e-mail: iaroslav.kolomiets@skoltech.ru}\equalcontrib
\and Miguel Altamirano Cabrera$^{1,2}$\thanks{e-mail: m.altamirano@skoltech.ru}\equalcontrib
\and Artem Lykov$^{2,1}$\thanks{e-mail: Artem.Lykov@skoltech.ru}\equalcontrib
\and Jeffrin Sam$^{1}$\thanks{e-mail: jeffrin.sam@skoltech.ru}
\and Dmitrii Iarchuk$^{1,2}$\thanks{e-mail: Dmitrii.Iarchuk@skoltech.ru}
\and Yara Mahmoud$^{1}$\thanks{e-mail: Yara.Mahmoud@skoltech.ru}
\and Daniia Zinniatullina$^{1,2}$\thanks{e-mail: daniia.zinniatulina@skoltech.ru}
\and Mikhail Konenkov$^{1,2}$\thanks{e-mail: mikhail.konenkov@skoltech.ru}
\and Dzmitry Tsetserukou$^{1}$\thanks{e-mail: d.tsetserukou@skoltech.ru}
}

\affiliation{
\scriptsize
$^{1}$Intelligent Space Robotics Lab, Skolkovo Institute of Science and Technology, Moscow, Russian Federation\\
$^{2}$R\&D Center, MWS, Moscow, Russian Federation\\
\equalcontribsymbol These authors contributed equally to this work.
}

\teaser{
  \vspace{-2.0em}
  \centering
  \includegraphics[width=0.7\linewidth]{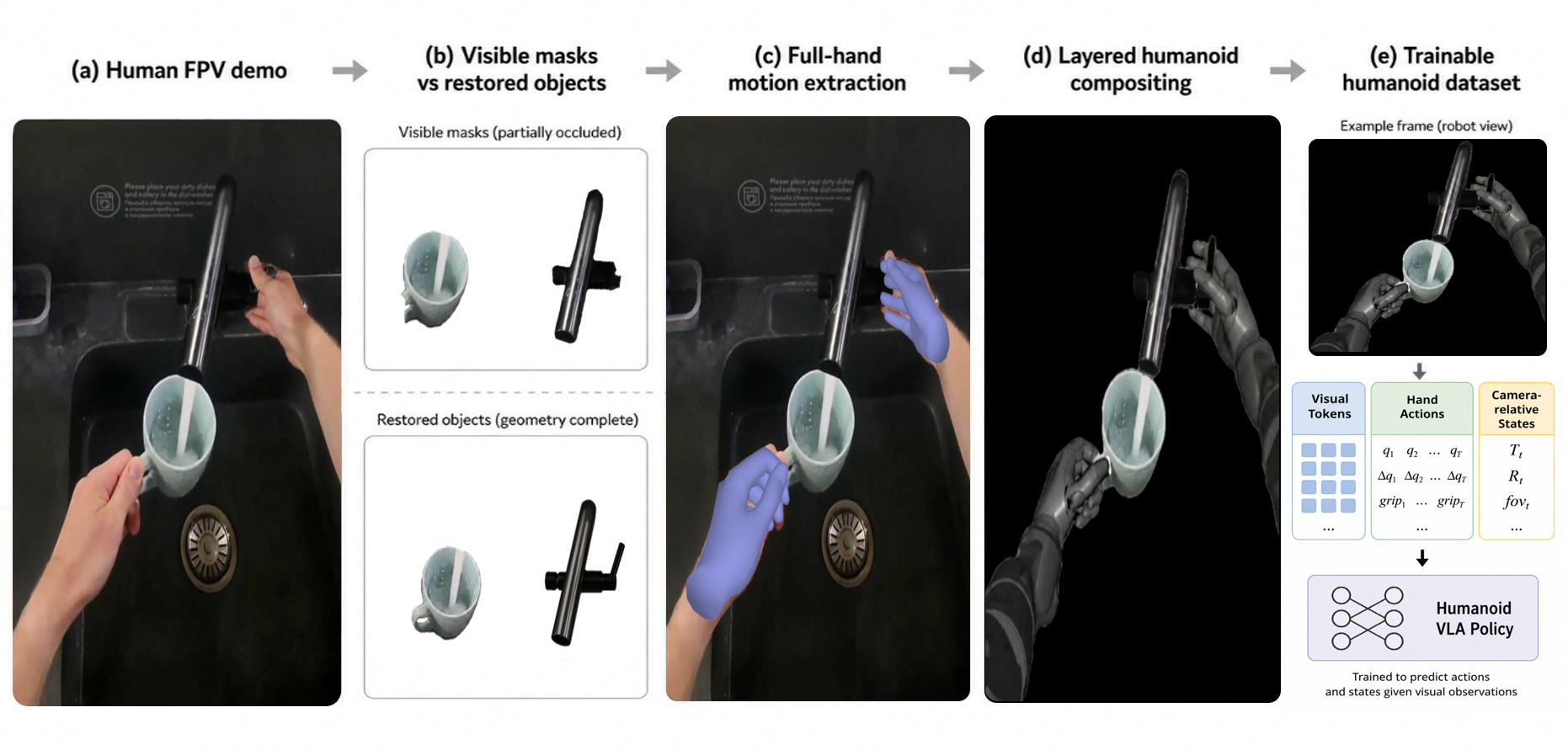}
  \vspace{-0.8em}
  \caption{AgenticFocus converts human FPV videos into robot-trainable mixed-reality demonstrations by restoring occluded objects, retargeting full-hand motion, and producing synchronized visual-action data.}
  \label{fig:teaser}
  \vspace{-1.0em}
}

\abstract{
Human egocentric video is a scalable supervision source for humanoid policy learning, but current pipelines struggle with hand-object occlusion, oversimplified motion, or specialized capture hardware. We introduce \textit{AgenticFocus}, a Mixed Reality synthesis pipeline that converts ordinary first-person-view human videos into robot-trainable demonstrations by restoring occluded object geometry, reconstructing full-hand motion, and retargeting it to a humanoid embodiment through camera-relative alignment and layered compositing. The resulting dataset pairs focused visual observations with synchronized robot actions and states. AgenticFocus achieves lower trajectory error and smoother wrist motion than cross-embodiment baselines, with SPARC scores of $-5.18$ versus $-5.56$ and $-6.05$.
}


\makeatletter
\renewcommand{\@fnsymbol}[1]{%
  \ensuremath{%
    \ifcase#1
    \or *
    \or \dagger
    \or \ddagger
    \or \mathsection
    \or \mathparagraph
    \or \|
    \or \#
    \or \dagger\dagger
    \or \ddagger\ddagger
    \or \mathsection\mathsection
    \else
      \@ctrerr
    \fi
  }%
}
\makeatother
\begin{document}


\firstsection{Introduction}
\setcounter{footnote}{1}
\maketitle

%% \section{Introduction} %for journal use above \firstsection{..} instead

Recent progress in Physical AI has made one limitation increasingly clear: the next frontier is not only better models but also better supervision. Across Vision-Language-Action (VLA) policies, world-action models, and related visuomotor architectures, performance ultimately depends on access to large amounts of robot-usable visual and action data~\cite{brohan2023rt2, kim2024openvla, wang2026worldactionmodels}. Yet for dexterous humanoids, such data remains exceptionally difficult to obtain. Teleoperation is costly, wearable capture systems are cumbersome, and embodiment-specific data collection does not scale across platforms~\cite{openx2023, wang2024dexcap, hoque2026egodex}. By contrast, human egocentric video is already abundant: it captures diverse object interactions in natural environments, often at internet scale, and therefore represents one of the most promising untapped supervision sources for humanoid manipulation~\cite{damen2020epic, hoque2026egodex, lepert2026masquerade}.

The challenge is that human video is not robot data. Converting a first-person human demonstration into a training signal for a humanoid policy requires bridging three coupled gaps. The first is a \textit{viewpoint gap}: human demonstrations are captured from head- or chest-mounted cameras, whereas humanoid observations are defined in robot-centered frames~\cite{lepert2026masquerade, liu2026egoengine, hoque2026egodex}. The second is an \textit{interaction-region gap}: during dexterous manipulation, the hand frequently occludes the manipulated object, precisely at the contact regions where geometry, boundaries, and local depth relationships are most important for control. Conventional actor removal, inpainting, or generative editing often fail at exactly these regions, corrupting the visual evidence that downstream policies most need~\cite{lepert2026masquerade, liu2026egoengine, paliwal2026doasido, li2026disentangled}. The third is an \textit{action grounding gap}: even when robot-like observations can be synthesized, they are often not paired with embodiment-consistent actions and states suitable for policy learning~\cite{liu2026egoengine, paliwal2026doasido, wang2024dexcap}.

A growing body of work has begun to address portions of this pipeline. \textit{DexCap} demonstrates that high-quality dexterous supervision can be collected with specialized motion-capture hardware~\cite{wang2024dexcap}. \textit{EgoDex} and \textit{EgoEngine} show the value of richer egocentric sensing setups for humanoid learning~\cite{hoque2026egodex, liu2026egoengine}. M\textit{asquerade, EgoEngine, }and related cross-embodiment video editing methods further establish that human demonstrations can be transformed into robot-like observations~\cite{lepert2026masquerade, liu2026egoengine, paliwal2026doasido, li2026disentangled}. However, these approaches still leave open a critical regime: ordinary monocular FPV video for dexterous humanoid learning. Some methods rely on dedicated capture hardware, stereo sensing, or scene-specific reconstruction~\cite{wang2024dexcap, hoque2026egodex, liu2026egoengine}. Others simplify human interaction to gripper-level abstractions, or degrade object geometry under heavy hand-object occlusion~\cite{lepert2026masquerade}. As a result, existing systems do not yet provide a robust path from ordinary human video to the focused visual-action supervision required by multi-fingered humanoid policies~\cite{liu2026egoengine, paliwal2026doasido, hoque2026egodex}.

This gap is especially consequential in modern agentic robot systems, where high-level reasoning and low-level control are often separated. A deliberative module identifies goals, relevant objects, and task context, while a reactive policy executes fine-grained motion~\cite{ahn2022saycan, huang2022innermonologue, driess2023palme}. In this setting, the controller does not require a complete reconstruction of the full scene; it requires a focused representation of task-relevant objects, local contact geometry, and embodiment-consistent motion. We therefore formulate human-to-humanoid transfer as Mixed Reality synthesis: preserve the physical interaction structure, replace the embodiment, and produce supervision in the form required by reactive control~\cite{milgram1994mixedreality, nechyporenko2024armada, mori2017diminishedreality, li2022e2fgvi}.

In this paper, we present \textit{AgenticFocus}, a structured Mixed Reality pipeline for converting ordinary human FPV video into robot-trainable demonstrations for dexterous humanoids. The central idea is to decouple scene semantics from embodiment. We first identify and track task-relevant objects across the sequence. Because visible masks alone are incomplete under hand occlusion, we restore stable object templates after human removal, preserving full object geometry rather than only the visible fragments. In parallel, we reconstruct full-hand human motion and retarget it through a camera-relative formulation that aligns the demonstration with the robot's virtual viewpoint. Finally, we compose the restored scene and retargeted robot through layered rendering, yielding visually stable interactions with plausible depth ordering between manipulated objects and robot fingers near contact.

By turning ordinary human video into robot-trainable supervision, AgenticFocus lowers the data barrier for humanoid manipulation in domains such as household and everyday-object handling, assistive and service robotics, and healthcare or eldercare support tasks. Because it removes the need for specialized capture hardware, the pipeline also serves as an accessible data source for robot-learning research and education. More broadly, it contributes to scalable, hardware-free collection of manipulation demonstrations for the embodied-AI community.

The result is not merely an edited video, but a synchronized representation that couples focused visual observations with robot actions and camera-relative states. This makes ordinary FPV human video a practical supervision source for humanoid visuomotor learning, while avoiding the capture burden of motion-capture gloves, stereo rigs, and scene-specific digital twins~\cite{wang2024dexcap, hoque2026egodex, liu2026egoengine}. Although the experiments in this paper focus on reactive humanoid control, the broader implication is more general: AgenticFocus addresses a central data-conversion bottleneck in Physical AI by transforming abundant human video into embodiment-consistent visual-action supervision~\cite{lepert2026masquerade, liu2026egoengine, paliwal2026doasido, openx2023, kim2024openvla}.

Our contributions are threefold. First, we introduce a focused Mixed Reality synthesis pipeline that converts ordinary human FPV video into robot-trainable demonstrations by combining task-relevant object selection, object restoration under occlusion, and humanoid mixed-reality compositing. Second, we propose full-hand cross-embodiment retargeting for dexterous humanoids, reconstructing and transferring full-hand human motion to a multi-fingered embodiment through camera-relative alignment, rather than collapsing demonstrations to simplified gripper-level interactions. Third, we design a layered, object-preserving compositing strategy for contact-rich manipulation that preserves manipulated object geometry and near-contact depth relationships, producing synchronized visual and action/state data suitable for training reactive humanoid visuomotor policies.

% Our contributions are summarized as follows:
% \vspace{-0.3em}
% \begin{itemize}
%     \setlength{\itemsep}{0.15em}
%     \setlength{\parskip}{0pt}
%     \setlength{\parsep}{0pt}
%     \setlength{\topsep}{0.2em}

%     \item \textit{Focused Mixed Reality synthesis for humanoid learning:} We introduce a pipeline that converts ordinary human FPV video into robot-trainable demonstrations by combining task-relevant object selection, object restoration under occlusion, and humanoid mixed-reality compositing.

%     \item \textit{Full-hand cross-embodiment retargeting for dexterous humanoids:} We reconstruct and transfer full-hand human motion to a multi-fingered humanoid embodiment through camera-relative alignment, rather than collapsing demonstrations to simplified gripper-level interactions.

%     \item \textit{Layered object-preserving compositing for contact-rich manipulation:} We propose a deterministic rendering strategy that preserves manipulated object geometry and near-contact depth relationships, producing synchronized visual and action/state data suitable for training reactive humanoid visuomotor policies.
% \end{itemize}
\vspace{-0.4em}

\section{Method}
\label{sec:method}

\subsection{Overview}
\begin{figure}[t]
\centering
\vspace{-0.4em}
\includegraphics[width=0.92\linewidth]{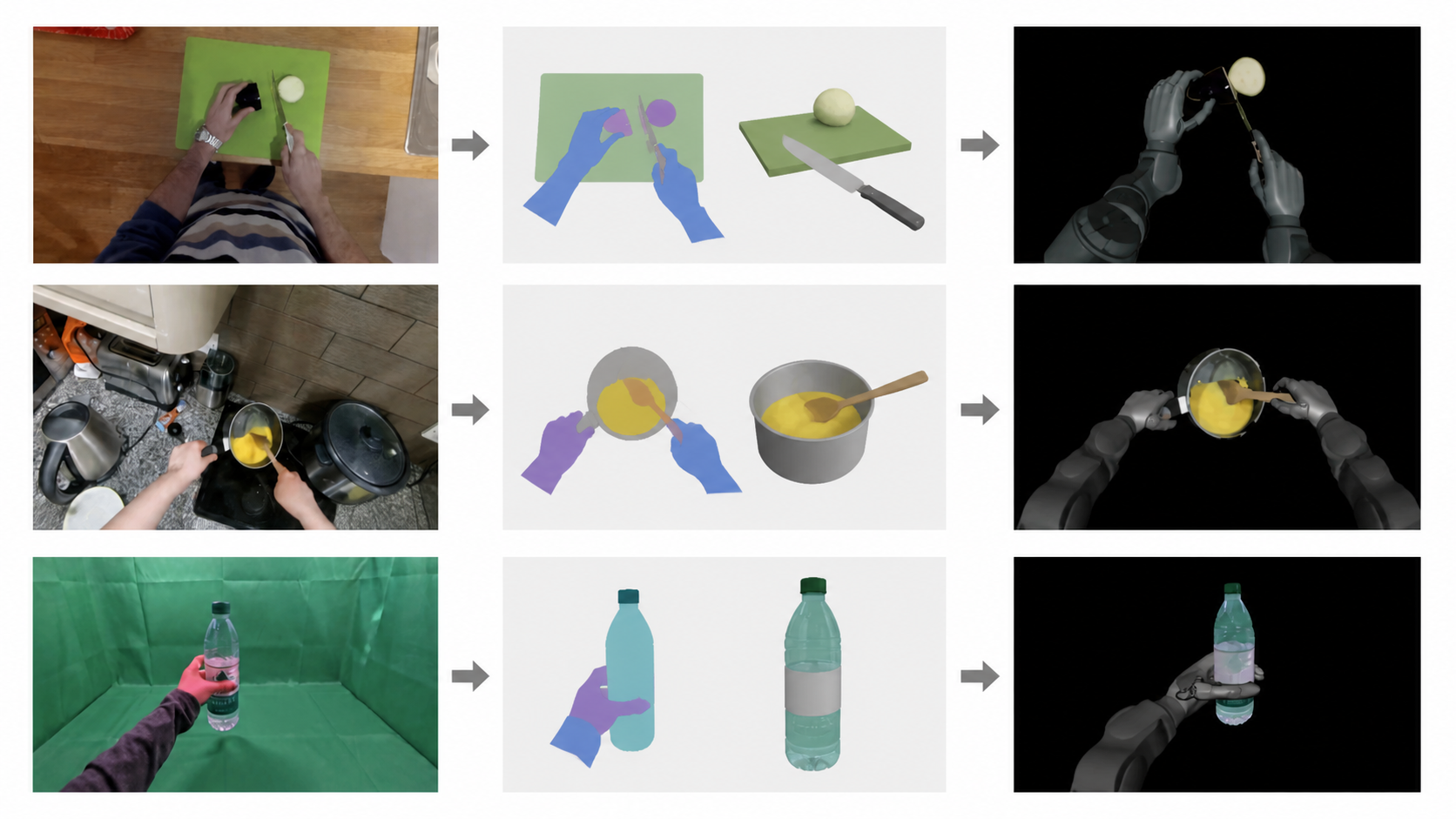}
\vspace{-0.2em}
\caption{Human-to-humanoid dataset conversion in AgenticFocus: human FPV manipulation, object-centric representation, and humanoid mixed-reality rendering.}
\label{fig:human_to_robot_examples}
\vspace{-0.6em}
\end{figure}

\textit{AgenticFocus} converts ordinary human FPV manipulation videos into synchronized humanoid training demonstrations consisting of (i) focused visual observations in which human hands are replaced by dexterous robot hands while task-relevant objects are preserved, and (ii) the corresponding robot action and state trajectories. The pipeline is guided by a simple principle: preserve scene semantics at the object level while transferring embodiment at the manipulator level, as illustrated in Fig.~\ref{fig:human_to_robot_examples}.

Given an input egocentric video, the method first identifies the task-relevant interaction object and tracks it across the sequence. It then removes the human actor while preserving object geometry under occlusion, reconstructs full-hand human motion, retargets it to a humanoid embodiment in a camera-relative frame, and synthesizes the final output through layered mixed-reality compositing. This decomposition avoids end-to-end generative translation and instead produces deterministic, robot-trainable visual-action pairs.

\subsection{Object-Centric Restoration}
\label{sec:object_restoration}
The first stage isolates the manipulated object from the full human video. A vision-language model selects the task-relevant object from the scene, after which a SAM-style segmentation model tracks the object mask across frames. This produces per-frame object masks, RGBA object crops, and debugging videos for mask verification. In practice, visible object masks alone are insufficient: when the human hand occludes the object, the observed mask contains only the visible fragments of the object rather than its full geometry.

To preserve the manipulated object during human removal, we build an object-preserving inpainting mask instead of directly erasing the full foreground. Let $\mathcal{M}_{\text{hand}}(t)$ denote the human hand or arm mask at frame $t$, and let $\mathcal{M}_{\text{obj}}(t)$ denote the target-object mask. We define the inpainting region as
\begin{equation}
\mathcal{M}_{\text{inpaint}}(t) = \mathcal{M}_{\text{hand}}(t) \cap \neg \mathcal{M}_{\text{obj}}(t),
\end{equation}
so that only actor pixels outside the protected object region are removed. We implement object and arm tracking with SAM2~\cite{ravi2024sam2}, and feed the resulting inpainting masks to E2FGVI-HQ~\cite{li2022e2fgvi} to reconstruct the background after actor removal.

Background inpainting alone is not sufficient for stable contact-rich synthesis. Even after the actor is removed, residual hand artifacts, partial object remnants, or boundary flicker may remain near the interaction region. We therefore restore a stable object template from a clean frame and reinsert it during final compositing. The template provides object appearance and geometry, while the per-frame object masks are used to track its spatial placement over time. This separation reduces ghosting, suppresses hand-induced boundary artifacts, and preserves full object geometry even when the source object is heavily occluded.

\subsection{Full-Hand Humanoid Retargeting}
\label{sec:retargeting}
In parallel, the method reconstructs human hand motion and transfers it to a dexterous humanoid embodiment. Human hand pose is estimated frame-by-frame using a 3D hand reconstruction model; in our implementation, finger and wrist motion are recovered using WiLoR/HaMeR-style modules~\cite{potamias2024wilor, pavlakos2024hamer}. The recovered hand topology is then mapped to the Unitree G1 arms and BrainCo dexterous hands through inverse kinematics in MuJoCo~\cite{todorov2012mujoco}.

A key challenge is that human egocentric motion is defined in the camera frame of the demonstrator, whereas the target robot must execute motion relative to its own embodiment and viewpoint. To bridge this gap, retargeting is formulated in a camera-relative coordinate system. Let $\mathbf{P}_{\text{human}}$ denote a 3D point in the human camera frame. We map it into the robot's virtual camera frame through
\begin{equation}
\mathbf{P}_{\text{robot}} = \beta \, \mathbf{R}_{\text{cam}} \mathbf{P}_{\text{human}} + \mathbf{T}_{\text{offset}},
\end{equation}
where $\mathbf{R}_{\text{cam}}$ is the camera-axis transformation, $\beta$ is the workspace scaling factor, and $\mathbf{T}_{\text{offset}}$ is the position of the virtual robot camera relative to the torso frame. This formulation decouples the demonstration from its original recording geometry and allows retargeting to be solved in an embodiment-consistent frame.

The resulting wrist and finger trajectories are smoothed before and during IK solving to improve temporal stability. For the robot joint configuration $\mathbf{q}_t$, we apply an exponential moving average
\begin{equation}
\mathbf{q}_{t} = \alpha \hat{\mathbf{q}}_{t} + (1-\alpha)\mathbf{q}_{t-1},
\end{equation}
where $\hat{\mathbf{q}}_{t}$ is the raw IK estimate and $\alpha$ is the smoothing coefficient. Fixed wrist-orientation corrections are additionally applied to account for the mounting geometry of the robot hands relative to the recovered human palm orientation. The output of this stage is a synchronized action-state representation containing arm trajectories, finger commands, and camera-relative robot states.

\subsection{Layered Mixed-Reality Compositing}
\label{sec:compositing}
The final stage synthesizes the robot demonstration by compositing the retargeted humanoid over the cleaned scene. A naive 2D overlay is insufficient for dexterous interaction: if the robot hand is rendered entirely in front of the object, contact looks implausible; if rendered entirely behind it, grasp structure is lost. This is especially problematic when fingers wrap around small manipulated objects.

To preserve local depth cues at the interaction region, we separate robot rendering into a full articulated-hand pass and a near-contact thumb pass. The restored object template is then composited around these renders so that the object can occlude the appropriate finger regions while preserving visible thumb contact where required by the grasp. This layered compositing strategy produces more plausible interaction structure than a single foreground overlay, particularly for small objects and wrap-around grasps.

In our implementation, rendering is performed in a headless MuJoCo environment using the Unitree G1 and BrainCo hand meshes derived from the robot description. Because the robot geometry is rendered from an explicit articulated model rather than generated by image synthesis, the final observation is visually stable and free of diffusion-style hallucination artifacts.

The final output of AgenticFocus is a paired dataset: a focused visual sequence in which human manipulators are replaced by humanoid hands while preserving the real task objects, together with synchronized robot actions and states suitable for downstream policy learning.

\section{Experiments}
\label{sec:experiments}

We evaluate AgenticFocus along two complementary dimensions of human-to-humanoid motion retargeting: spatial accuracy with respect to ground-truth trajectories and temporal smoothness of the retargeted humanoid wrist motion.

\subsection{Experimental Setup}
We construct evaluation sequences from egocentric human videos, including clips from EPIC-KITCHENS~\cite{damen2020epic} and internally processed demonstrations. All videos are standardized to 30 FPS. For each clip, the pipeline produces two synchronized outputs: (i) a rendered mixed-reality video in which human hands are replaced by Unitree G1 and BrainCo hands while the target object remains in the real scene, and (ii) a structural action-state record stored as \texttt{.npz}/\texttt{.json} containing arm trajectories, finger commands, and camera-relative robot states.

We compare AgenticFocus against Masquerade~\cite{lepert2026masquerade} and Do as I Do~\cite{paliwal2026doasido} under the same evaluation protocol. Before comparison, all trajectories are temporally resampled to a common normalized timeline and expressed in the same coordinate frame.

\subsection{Trajectory Accuracy}
\label{sec:exp_accuracy}

\paragraph{Metric.}
We compute the mean 3D position error between the ground-truth trajectory $\mathbf{p}_t$ and the reconstructed trajectory $\hat{\mathbf{p}}_t$ produced by each method:
\begin{equation}
e_t = 100 \left\| \hat{\mathbf{p}}_t - \mathbf{p}_t \right\|_2
= 100 \sqrt{
(\hat{x}_t-x_t)^2+
(\hat{y}_t-y_t)^2+
(\hat{z}_t-z_t)^2 },
\end{equation}
where the factor 100 converts meters to centimeters. The final error is averaged over all frames and evaluation clips; error bars denote the 95\% confidence interval across episodes.

\begin{figure}[!h]
    \centering
    \includegraphics[width=0.8\linewidth]{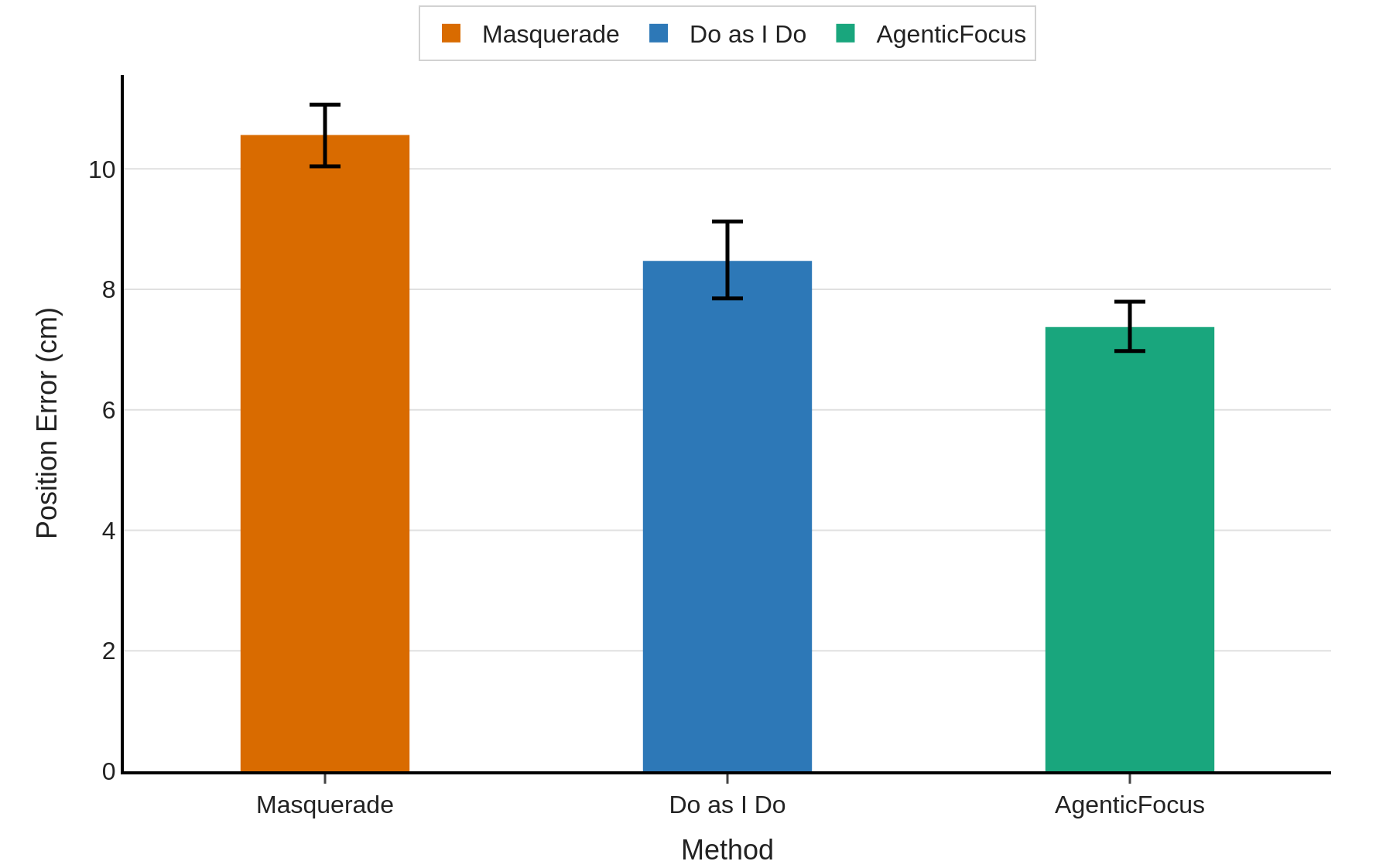}
    \vspace{-0.5em}
    \caption{
    Mean 3D position error for trajectory reconstruction.
    Lower values indicate better agreement with the ground-truth trajectory.
    Error bars show the 95\% confidence interval.
    }
    \label{fig:mean_position_error}
    \vspace{-0.8em}
\end{figure}

\paragraph{Results.}
Fig.~\ref{fig:mean_position_error} reports the mean 3D position error across evaluation episodes. AgenticFocus achieves the lowest reconstruction error among the compared methods, indicating better agreement with the ground-truth trajectory. This suggests that camera-relative alignment and full-hand retargeting reduce spatial mismatch compared with the baseline human-to-robot transfer pipelines.

\subsection{Retargeting Smoothness}
\label{sec:exp_smoothness}

\begin{figure}[!t]
\centering

\begin{subfigure}[t]{\linewidth}
\centering
\includegraphics[width=0.75\linewidth]{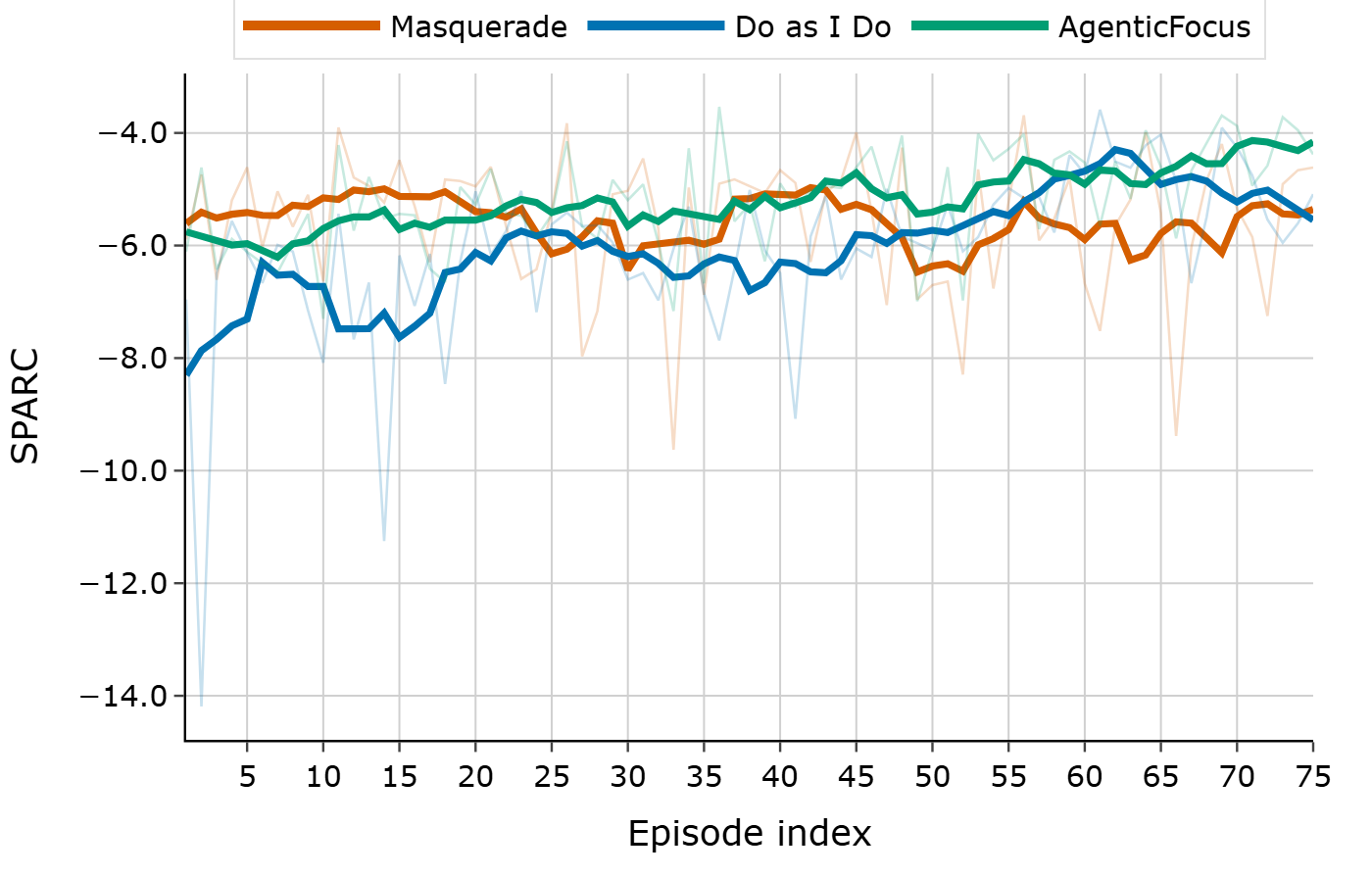}
\caption{Episode-wise SPARC across validation episodes. Solid curves show a 7-episode rolling mean.}
\label{fig:per_episode_sparc_3methods}
\end{subfigure}

\vspace{0.6em}

\begin{subfigure}[t]{0.75\linewidth}
\centering
\includegraphics[width=\linewidth]{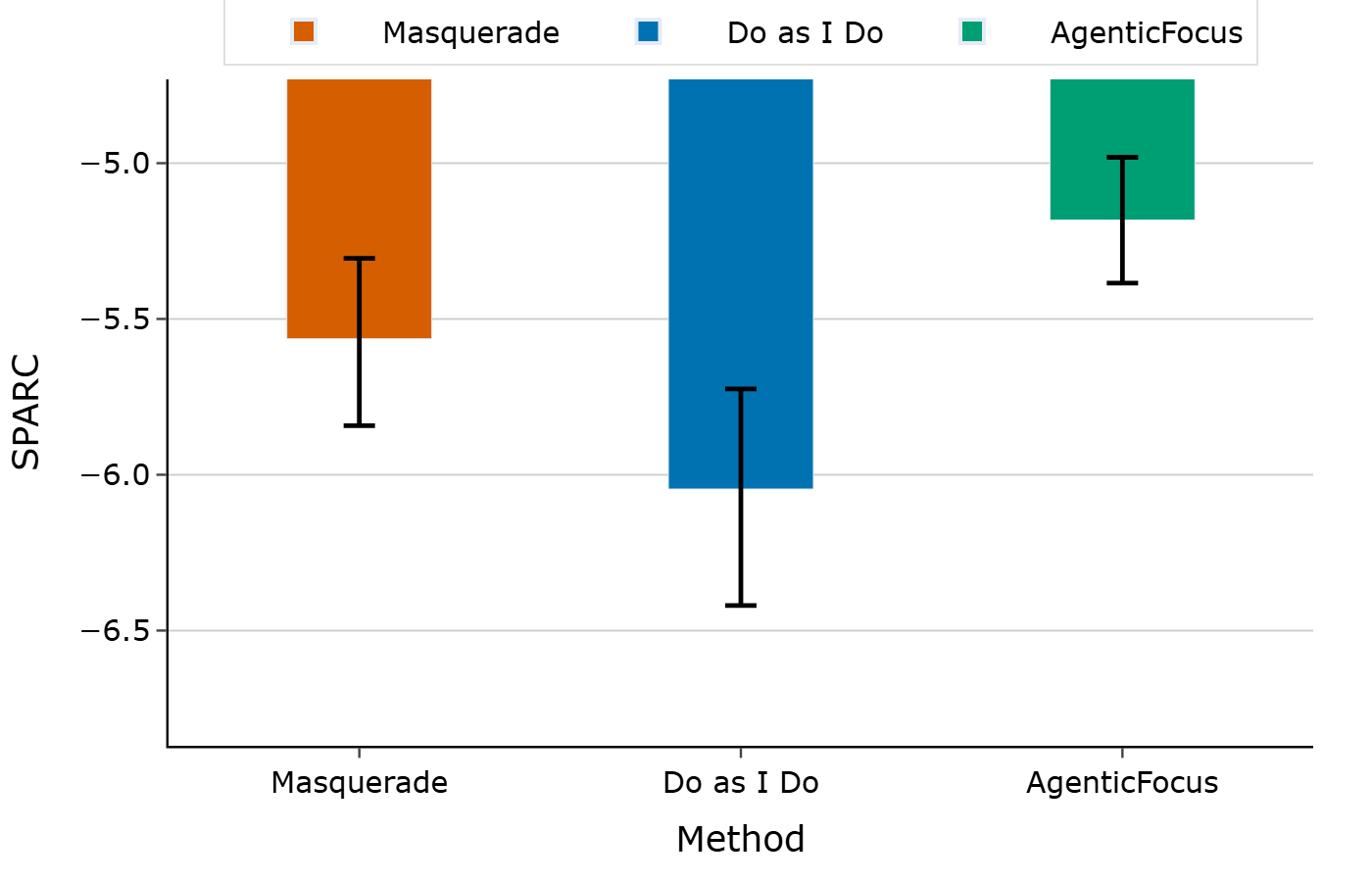}
\caption{Mean SPARC with 95\% bootstrap confidence intervals over 75 episodes. Less negative values correspond to smoother motion.}
\label{fig:mean_sparc_ci_3methods}
\end{subfigure}

\caption{
SPARC-based trajectory smoothness comparison across Masquerade, Do as I Do, and AgenticFocus.
}
\label{fig:sparc_smoothness_comparison}
\end{figure}

\paragraph{Metric.}
We quantify temporal smoothness using the Spectral Arc Length (SPARC) metric~\cite{balasubramanian2012robust}, computed from the humanoid wrist speed profile. Let $V(\omega)$ denote the Fourier magnitude spectrum of the wrist speed signal, and let $\hat{V}(\omega)=V(\omega)/V(0)$ be the normalized spectrum. SPARC is defined as
\begin{equation}
\mathrm{SPARC} =
-\int_{0}^{\omega_c}
\sqrt{
\left(\frac{1}{\omega_c}\right)^2+
\left(\frac{d\hat{V}(\omega)}{d\omega}\right)^2
}
\,d\omega ,
\end{equation}
where $\omega_c$ is the cutoff frequency. SPARC values are negative by construction; less negative values indicate smoother and more temporally stable motion.

\paragraph{Results.}
Fig~\ref{fig:sparc_smoothness_comparison}(a) plots episode-wise SPARC across 75 validation episodes. AgenticFocus consistently maintains the least negative SPARC throughout, indicating smoother and more coherent retargeted wrist motion than both baselines. Fig~\ref{fig:sparc_smoothness_comparison}(b) reports mean SPARC with 95\% bootstrap confidence intervals: AgenticFocus attains $-5.18$ (95\% CI $[-5.38,\,-4.98]$), compared to $-5.56$ ($[-5.84,\,-5.31]$) for Masquerade and $-6.05$ ($[-6.42,\,-5.72]$) for Do as I Do. Relative to Masquerade, AgenticFocus reduces the SPARC magnitude by approximately 7\%, and by roughly 14\% relative to Do as I Do. These results suggest that camera-relative retargeting and dedicated trajectory smoothing substantially improve robot-ready humanoid wrist motion.

% \subsection{Downstream Robot Learning}
% \label{sec:exp_policy}
% The final experiment evaluates whether the synthesized demonstrations can be used to train a real humanoid visuomotor policy. The target platform is a Unitree G1 humanoid equipped with BrainCo dexterous hands, and the generated visual-action pairs are used as supervision for reactive manipulation learning.

% The initial downstream evaluation is performed on real-robot object manipulation tasks, including pick-and-place settings in which success can be measured over repeated trials. The analysis examines whether policies trained on AgenticFocus data can execute the intended manipulation behavior and whether the proposed representation provides a practical supervision source for physical deployment.

\section{Conclusion}
\label{sec:conclusion}
This paper presented AgenticFocus, a structured mixed-reality pipeline for converting ordinary human FPV videos into synchronized humanoid training demonstrations. By combining object-centric restoration, camera-relative full-hand retargeting, and layered compositing, the method preserves task-relevant object structure while transferring dexterous manipulation into a robot-consistent visual-action representation. Empirically, AgenticFocus achieves lower trajectory error and smoother retargeted wrist motion (SPARC) than existing cross-embodiment baselines, indicating that camera-relative alignment and dedicated trajectory smoothing meaningfully improve the fidelity of synthesized humanoid demonstrations.

The resulting dataset couples focused visual observations with robot actions and camera-relative states, providing a practical supervision source for humanoid visuomotor learning without relying on specialized capture hardware or fully generative scene translation. Extending this evaluation to downstream policy training and broader embodiments is a natural direction for future work. More broadly, the method addresses a central bottleneck in cross-embodiment data conversion: transforming abundant egocentric human video into robot-trainable demonstrations at scale.

\bibliographystyle{abbrv-doi}

\bibliography{bib}
\end{document}